\documentclass[3p,times]{elsarticle}

\begin{document}
\begin{frontmatter}




\title{A General Framework for Multi-focal Image Classification and Authentication: Application to Microscope Pollen Images}


\author[insp]{Fran\c{c}ois Chung\corref{corr}}
\ead{francois.chung@gmail.com}
\cortext[corr]{Corresponding author}
\author[insp]{Tom\'{a}s Rodr\'{i}guez}

\address[insp]{Inspiralia, Calle La Estrada 10B, 28034 Madrid, Spain}

\begin{abstract}
In this article, we propose a general framework for multi-focal image classification and authentication, the methodology being demonstrated on microscope pollen images. The framework is meant to be generic and based on a brute force-like approach aimed to be efficient not only on any kind, and any number, of pollen images (regardless of the pollen type), but also on any kind of multi-focal images. All stages of the framework's pipeline are designed to be used in an automatic fashion. First, the optimal focus is selected using the absolute gradient method. Then, pollen grains are extracted using a coarse-to-fine approach involving both clustering and morphological techniques (coarse stage), and a snake-based segmentation (fine stage). Finally, features are extracted and selected using a generalized approach, and their classification is tested with four classifiers: Weighted Neighbor Distance, Neural Network, Decision Tree and Random Forest. The latter method, which has shown the best and more robust classification accuracy results (above 97\% for any number of pollen types), is finally used for the authentication stage.
\end{abstract}


\end{frontmatter}


\section{Introduction}

Bee products are known to have important nutritive and curative properties \cite{Saa-Otero2000}. However, these properties are not currently properly guaranteed in Europe, as there are no standards at European level for certain bee products like pollen and royal jelly. This means that it is possible to find products in the market under these labels without any quality and authenticity control. As a consequence, there has been a significant interest of the scientific community for the study and recognition of bee-related products such as honey, royal jelly and honeybee pollen.

Pollen is collected by bees in the form of ball-shaped loads known as \emph{pollen loads}. Studies have shown that these loads are \textit{monospecific}, meaning they are composed of grains extracted from the same plant taxon \cite{Corrion2004}. Pollen grains have specific morphological and textural properties that vary from one pollen type, or \emph{taxon}, to another. These properties referred to as \textit{features} include, among others, the size, shape, color and texture of the grain and are used as discriminant properties for the classification of pollen from different species.

This classification is performed by expert palynologists not only to study their nutritional and therapeutical properties, but also to characterize their floral and geographical origin. The process of manually separating pollen loads using the mere color information has been and is still widely used by palynologists to classify loads by pollen types. However, this process is time-consuming, subjective and requires highly trained palynologists. These issues have been acknowledged by palynologists in the literature and many methods to automate the classification process have been proposed since then.

The main contribution of these methods resides in the use of image processing for the automatic classification of pollen types \cite{DelPozo2012}. For instance, pollen loads have been classified using color information extracted from camera images \cite{Chica2012}. However, this macroscopic color-based system is not accurate enough to robustly deal with numerous pollen types, should they feature a similar color. Nonetheless, this procedure can be applied as a pre-processing step prior to more robust and accurate methods, such as those using microscope images \cite{Wu2008}.

The Scanning Electron Microscope (SEM) \cite{Langford1990} was the first relevant work involving microscope images. Pollen grains were characterized by texture features such as the co-occurrence matrix. Although considered as successful, this system has also been reported as expensive, slow and difficult to implement in a daily routine. The Confocal Laser Scanning Microscope has also been proposed \cite{Ronneberger2002}, but for the same reasons, such a hardware solution does not seem suitable for a practical use. So far, the preferred solution for acquiring pollen images is the Light Microscope (LM).

In LM-based applications, discriminative features \cite{Gurevich2006}, which are extracted from the acquired microscope images, are selected to be representative of the different pollen types. These features are often based on texture and shape, and may be extracted after a multi-scale filtering scheme \cite{Corrion2004}. Texture-based features are usually extracted using sub-images of pollen grains separately cropped from the microscope image. Shape-based features, such as area, perimeter, diameter, roundness and thickness, may also be added to reinforce the discrimination between pollen types \cite{Rodriguez2004}. However, in this case, the feature extraction must be performed on binary images, or \textit{masks}, separating the pollen grain (foreground) from the rest of the image (background). More sophisticated shape features, such as Fourier descriptors \cite{Rodriguez2004} and moment invariants \cite{Zhang2004}, have been proposed to improve the discriminative power of morphological features.

In the literature, several methods have been implemented for the feature-based classification of pollen images. The Minimum Distance Classifier (MDC) combined with the Mahalanobis distance has shown to perform well \cite{Rodriguez2004}. Same classifier has been compared to both Support Vector Machine (SVM) and Multilayer Perceptron Neural Network (MLP) \cite{Rodriguez2006}. Neural Networks (NNs) have been the subject of much attention from the scientific community. Their first implementation for pollen identification consisted in a feed-forward neural network with Haralick texture features as input data \cite{Li1999}. A notable work combines the pollen grain detection and classification, both performed using a NN named Pattern Recognition Architecture for Deformation Invariant Shape Encoding (Paradise) \cite{France2000}.

As depicted in the title, the overall objective of this article is to propose a general framework for multi-focal image classification and authentication, the methodology being demonstrated on microscope pollen images. Unlike other methods proposed in the literature, where discriminant features are chosen with respect to specific pollen types, our framework is based on a brute force-like approach that is not tailored to any specific number, or kind, of pollen types. Overall, the approach is meant to be generic, so as to perform on any kind of multi-focal images \cite{Landsmeer2009}. The idea is to first extract a large set of features (to account for the largest possible number of pollen properties), and then to filter them with respect to a given training dataset (to optimize computation time).

\begin{figure}
 \begin{center}
  \includegraphics[scale=0.45]{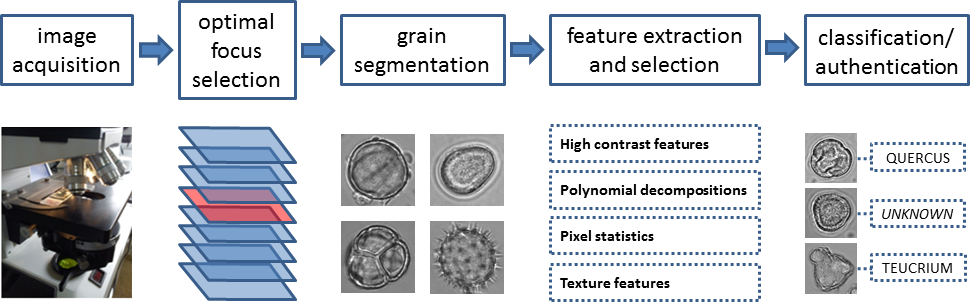}\\
 \end{center}
 \caption{Consecutive steps of our pipeline-based framework, from microscope image acquisition to pollen classification/authentication.}
 \label{fig-pipeline}
\end{figure}

The consecutive steps of our pipeline-based framework are depicted in Figure \ref{fig-pipeline}. First, multi-focal microscope images are acquired (Section \ref{sec-material}). In this work, 15 pollen types have been used to test the classification accuracy of our framework (Section \ref{sec-data}). The optimal focal image is selected (Section \ref{sec-foc-selection}) and then segmented to extract sub-images of pollen grains (Section \ref{sec-grain-seg}). Finally, features, which are extracted and selected from these sub-images (Section \ref{sec-features}), are used to classify (Section \ref{sec-classification}), or authenticate (Section \ref{sec-authentication}), the pollen grains. The authentication accuracy of our framework is tested with 7 additional pollen types, considered as outliers. A quite similar framework designed for the automatic recognition of biological particles has been proposed in the literature \cite{Ranzato2007}. However, this framework does not deal with multi-focal images, nor with color information, and does not consider pollen authentication. Furthermore, the larger number of extracted features on which is based our framework, i.e. 2164 (color images) or 1025 (grayscale images) \emph{vs.} 108 (only grayscale images), ensures a more varied pool of features aiming at improving the discrimination between pollen types.

\section{Material}
\label{sec-material}

To test our general framework, we used the Nikon Eclipse E200-LED bright-field microscope featuring 10~$\times$, 20~$\times$ and 40~$\times$ objectives. The microscope is coupled with the Nikon Digital Sight DS-Fi1 high resolution camera, which acquires the microscope pollen images and transfers them to a computer through a USB connection. This microscope camera is a 5-megapixel charge-coupled device (CCD) capturing color images at 2560$\times$1920 pixel resolution.

Prior to the image acquisition, pollen grains must be extracted from pollen loads and placed on a microscope slide. This extraction is performed using ethyl alcohol to clean the slide, silicone grease (or glycerine) to collect pollen grains, and a forceps to handle the slides. After the extraction of pollen grains, slides are dried with a heater.

\begin{figure}
 \begin{center}
  $\begin{array}{ccc}
   \multicolumn{1}{l}{} & \multicolumn{1}{l}{} & \multicolumn{1}{l}{} \\
   \includegraphics[scale=0.12]{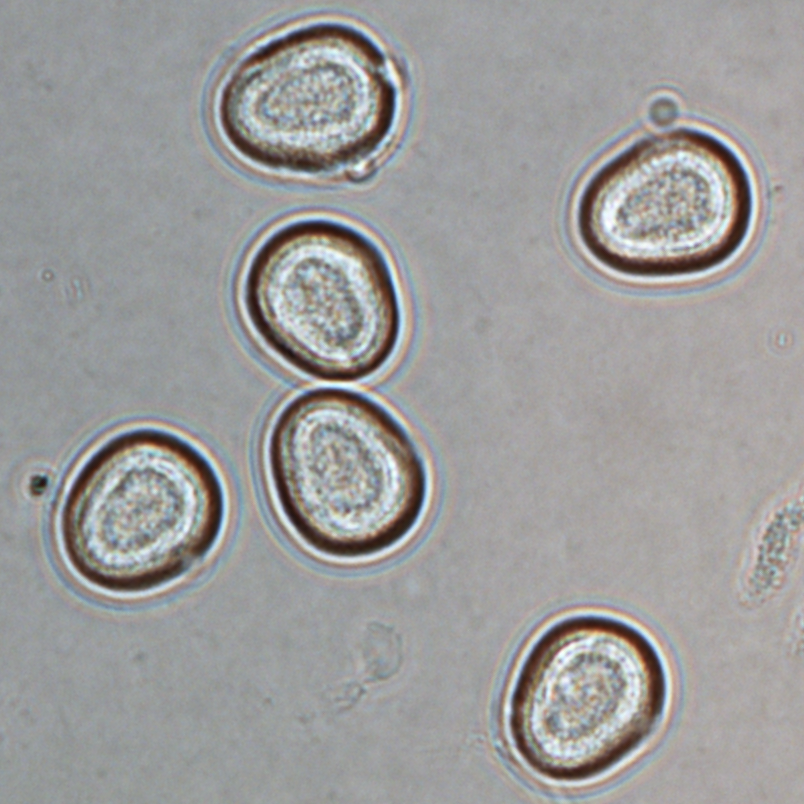} &
   \includegraphics[scale=0.12]{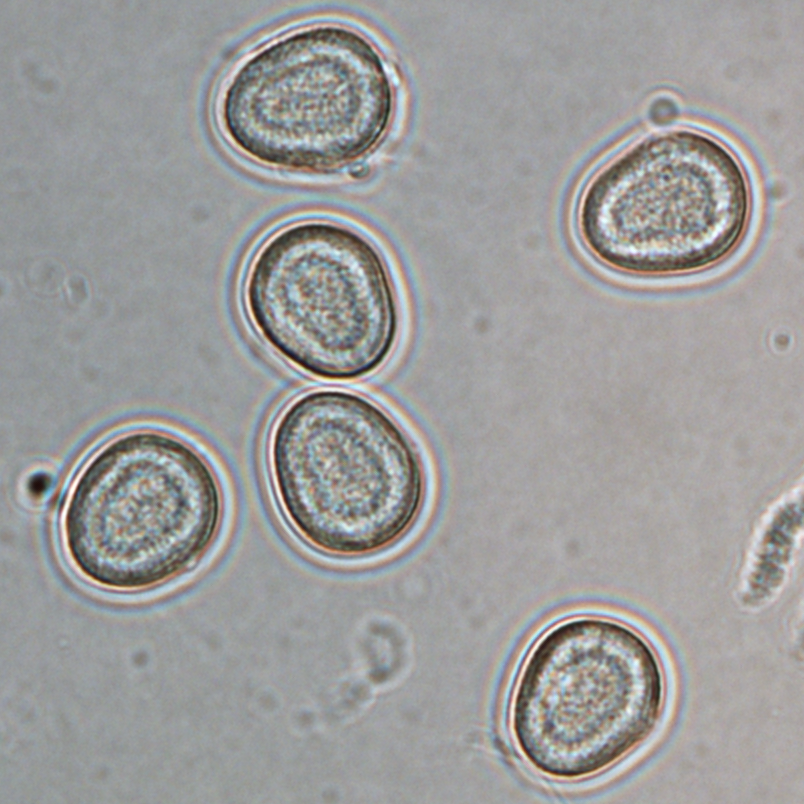} &
   \includegraphics[scale=0.12]{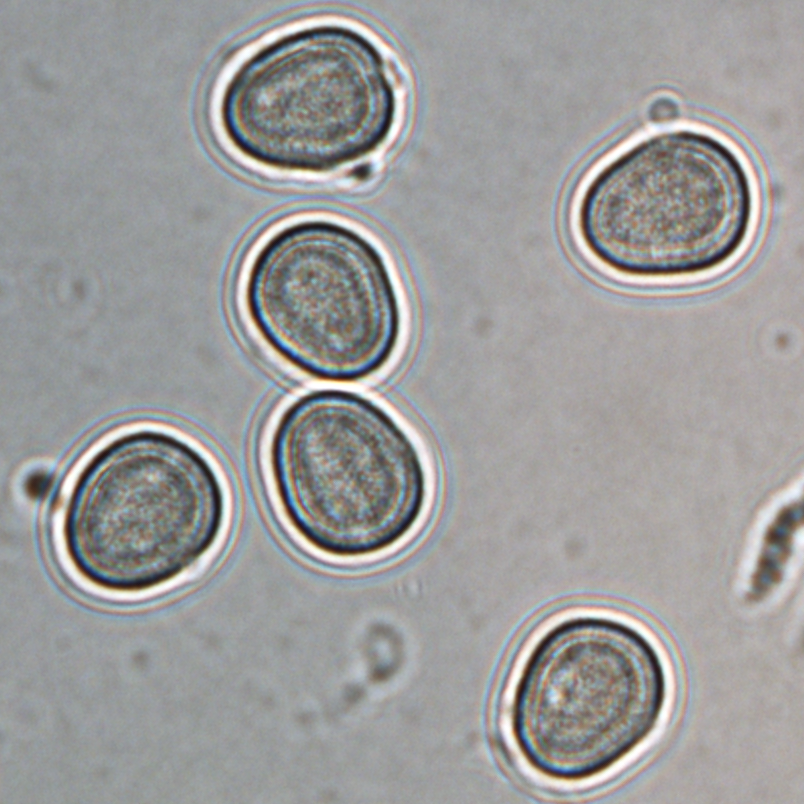} \\
   \mbox{(a)} & \mbox{(b)} & \mbox{(c)}
  \end{array}$
 \end{center}
\caption{Three focal planes (a,b,c) from a microscope image of pollen grains. Pollen type is \emph{Echium}. Each focus highlights a specific part of pollen grains, from the inner part to the exine.}
\label{fig-focal_planes}
\end{figure}

Multiple focal planes are acquired to highlight various parts of pollen grains, from the inner part to the exine. Each focal plane provides the system with useful textural and morphological information, which is used to discriminate one pollen type from another. An example of three focal planes is given in Figure \ref{fig-focal_planes}. For each microscope image, 31 focal planes are acquired, one from the equatorial plane, which is located at the center of the pollen grain \cite{Hesse2009}, and 15 in both upward and downward directions using a step of 1~$\mu m$/frame.

\section{Data}
\label{sec-data}

\begin{figure}
 \begin{center}
  $\begin{array}{ccccc}
   \multicolumn{1}{l}{} & \multicolumn{1}{l}{} & \multicolumn{1}{l}{} & \multicolumn{1}{l}{} & \multicolumn{1}{l}{}\\
   \includegraphics[scale=0.22]{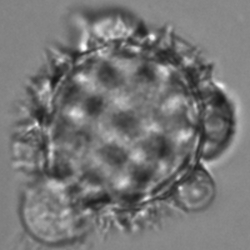} & \includegraphics[scale=0.22]{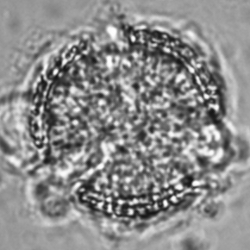} & \includegraphics[scale=0.22]{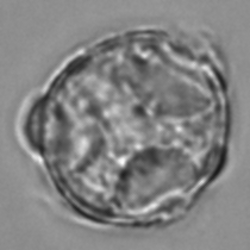} & \includegraphics[scale=0.22]{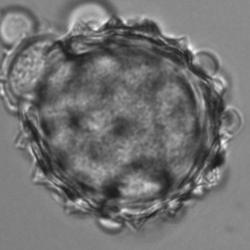} & \includegraphics[scale=0.22]{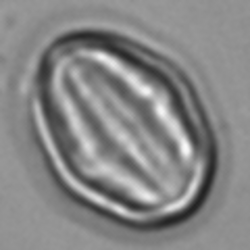}\\
   \mbox{(a)} & \mbox{(b)} & \mbox{(c)} & \mbox{(d)} & \mbox{(e)} \\[2ex]
   \includegraphics[scale=0.22]{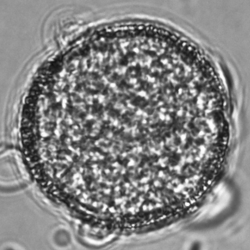} & \includegraphics[scale=0.22]{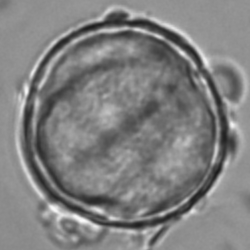} & \includegraphics[scale=0.22]{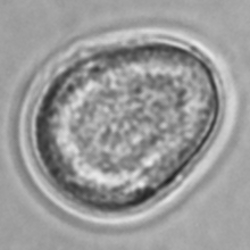} & \includegraphics[scale=0.22]{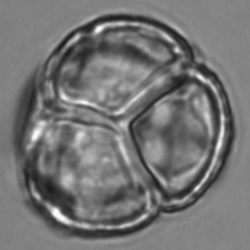} & \includegraphics[scale=0.22]{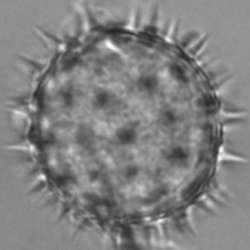}\\
   \mbox{(f)} & \mbox{(g)} & \mbox{(h)} & \mbox{(i)} & \mbox{(j)} \\[2ex]
   \includegraphics[scale=0.22]{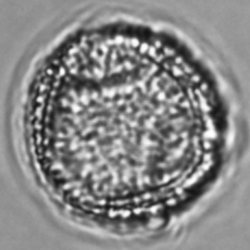} & \includegraphics[scale=0.22]{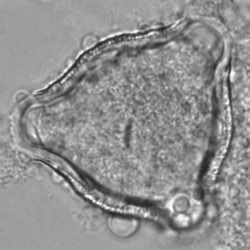} & \includegraphics[scale=0.22]{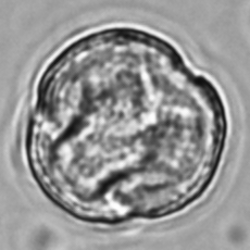} & \includegraphics[scale=0.22]{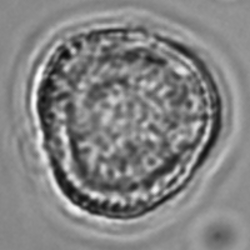} & \includegraphics[scale=0.22]{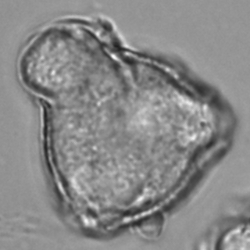}\\
   \mbox{(k)} & \mbox{(l)} & \mbox{(m)} & \mbox{(n)} & \mbox{(o)} \\[2ex]
  \end{array}$
 \end{center}
\caption{Bright-field microscope sub-images of pollen grains belonging to: (a) \emph{Aster}, (b) \emph{Brassica}, (c) \emph{Campanulaceae}, (d) \emph{Carduus}, (e) \emph{Castanea}, (f) \emph{Cistus}, (g) \emph{Cytisus}, (h) \emph{Echium}, (i) \emph{Ericaceae}, (j) \emph{Helianthus}, (k) \emph{Olea}, (l) \emph{Prunus}, (m) \emph{Quercus}, (n) \emph{Salix}, and (o) \emph{Teucrium} pollen types.}
\label{fig-pollen_types}
\end{figure}

The number of possible pollen types, or \emph{taxa}, in each country is high. However, there exists a few number of dominant types, which make them suitable candidates to test the classification accuracy of our framework. As this work aims to be generic, selected pollen types originate from multiple countries. Dominant pollen types from Spain, Italy and Turkey have been selected to validate our framework: \emph{Aster}, \emph{Brassica}, \emph{Campanulaceae}, \emph{Carduus}, \emph{Castanea}, \emph{Cistus}, \emph{Cytisus}, \emph{Echium}, \emph{Ericaceae}, \emph{Helianthus}, \emph{Olea}, \emph{Prunus}, \emph{Quercus}, \emph{Salix} and \emph{Teucrium}. In total, we have a pollen image database comprising 15 pollen types. A brief description of each pollen type is given in Table \ref{tab-type-overview} and some bright-field microscope sub-images are shown in Figure \ref{fig-pollen_types}.

\begin{table}[t]
\centering
\begin{tabular}{llll}
Pollen type & Size ($\mu m$)& Shape                 & Origin\\
\hline
\hline
Aster           & 20-40     & tri-tetracolporate    & Italy \\
\hline
Brassica        & $\sim25$  & tricolpate            & Spain \\
\hline
Campanulaceae   & 15-30     & tetraporate           & Spain \\
\hline
Carduus         & 20-50     & tricolporate          & Spain \\
\hline
Castanea        & 8-16      & tricolporate          & Italy/Spain\\
\hline
Cistus          & 26-50     & spheroidal            & Spain\\
\hline
Cytisus         & 15-30     & tricolporate          & Spain\\
\hline
Echium          & 10-25     & prolate               & Spain\\
\hline
Ericaceae       & 15-25     & tricolporate          & Spain\\
\hline
Helianthus      & 20-40     & tricolporate          & Bulgary\\
\hline
Olea            & 10-25     & spheroidal            & Spain\\
\hline
Prunus          & 26-50     & spheroidal            & Spain\\
                &           & tricolporate          &       \\
\hline
Quercus         & 26-50     & spheroidal            & Spain\\
\hline
Salix           & 16-30     & tricolporate          & Spain\\
                &           & prolate               &       \\
\hline
Teucrium        & 30-40     & tricolporate          & Turkey\\
\hline
\end{tabular}
\caption{Brief description of the pollen types selected to test the classification accuracy of our framework.}
\label{tab-type-overview}
\end{table}

\section{Optimal focus selection}
\label{sec-foc-selection}

The multi-focal acquisition of a microscope image ends up with the creation of a stack of consecutive focal planes. Extracting textural and morphological information from the entire stack is time-consuming and likely to generate irrelevant information, such as non-discriminative features, or noise. The most efficient and straightforward way to deal with this amount of information consists in selecting an optimal focal plane from the stack and consider it as the most representative. This process is referred to as \emph{auto-focusing} and is best known as the common autofocus function featured by most electronic cameras. In this case, the selection of the optimal focus is automatic, embedded, and consists in computing a focus measure in images acquired at several lens positions (i.e. consecutive focal planes) and in moving the lens to the position where the measure is a maximum \cite{Redondo2011}.

Regarding microscope images, the focus selection may be performed as a post-processing step after image acquisition. In the literature, this process is usually manual and performed by experts who carefully select the optimal focal image with respect to specific features needed to be highlighted. For pollen images, the focus is usually selected so as to highlight specific regions of the pollen grain, such as the exine or the inner part, in an attempt to maximize the discrimination between one pollen type from another. This process is time-consuming and not suitable for a daily routine. Furthermore, this process is not appropriate for our generic approach aiming to deal with any random pollen type.

\begin{figure}
 \begin{center}
  \includegraphics[scale=0.45]{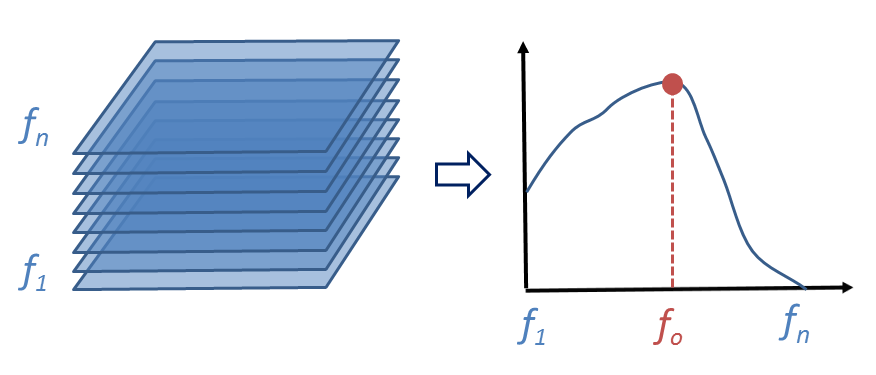}\\
 \end{center}
 \caption{To select the optimal focus $f_o$, the same image is acquired at consecutive focal planes (from $f_1$ to $f_n$, see left) and then, a criterion applied on the focal plane images is maximized (see right).}
 \label{fig-optimal-focus}
\end{figure}

This is why focus selection is a core component of our framework, as the selection of the best focus ensures the extraction of optimal features for the classification step. Automatic methods for the selection of the optimal focus have a similar approach as auto-focusing. The same image is acquired at consecutive focal planes and a criterion applied on the focal plane images is maximized (see Figure \ref{fig-optimal-focus}). A study on focus measures applied on bright-field microscope images for tuberculosis detection \cite{Osibote2010} showed that Vollath's $F_4$ measure gives the best results. Six methods defined in the spatial domain were tested and compared with respect to accuracy, execution time, range, full width at half maximum of the peak, and the presence of local maxima.

All methods for selecting the optimal focus are mainly optimized to be both visually and time-efficient. In the literature, these methods are mostly based on the derivative, statistics and histograms. In the first case, the best focus image is considered as the image with the highest intensity differences at the edges, e.g. using the gradient \cite{Redondo2011} or wavelets \cite{Yang2003}, in the hypothesis that a sharper edge means a better focused image. In the second case, the selection of the best focus image is performed by computing statistics on mathematical functions, such as correlation and variance. Finally, histogram-based methods extract measures from histogram analysis, such as range and entropy. A presentation and analyze of optimal focus methods may be found in \cite{Redondo2011}.

\section{Grain extraction}
\label{sec-grain-seg}

Pollen grain segmentation consists in extracting sub-images of pollen grains from microscope images (i.e. one pollen grain per sub-image). In the literature, sub-image extraction is performed through segmentation, whether manual, semi-automatic, or automatic \cite{France2000}. In this work, we propose an automatic segmentation procedure based on a coarse-to-fine approach.

In addition to the sub-image extraction, which consists in cropping the microscope image at pollen grain level (see Figure \ref{fig-grain-seg-fine}a), our method extracts the binary image, or \textit{mask}, of each grain sub-image (see Figure \ref{fig-grain-seg-fine}d). In the literature, mask extraction is usually not taken into account as the background of microscope images is considered as homogeneous (i.e. featuring a uniform color or texture), which rather limits its influence on both segmentation and classification results.

\begin{figure}
 \begin{center}
  $\begin{array}{ccccc}
   \multicolumn{1}{l}{} & \multicolumn{1}{l}{} & \multicolumn{1}{l}{} & \multicolumn{1}{l}{} & \multicolumn{1}{l}{}\\
   \includegraphics[scale=0.24]{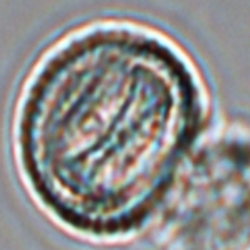} & \includegraphics[scale=0.24]{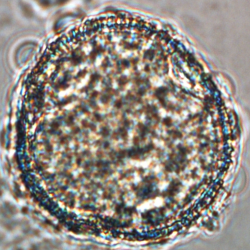} & \includegraphics[scale=0.24]{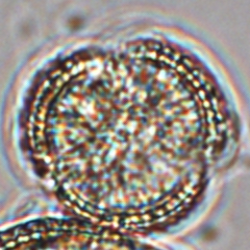} & \includegraphics[scale=0.24]{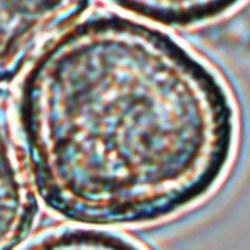} & \includegraphics[scale=0.24]{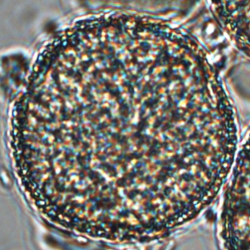}\\
   \mbox{(a)} & \mbox{(b)} & \mbox{(c)} & \mbox{(d)} & \mbox{(e)} \\
  \end{array}$
 \end{center}
\caption{Microscope sub-images of pollen grains whose non-homogeneous background features: (a,b) debris, (c,d) grain clusters, and (e) both.}
\label{fig-grain-seg-intro}
\end{figure}

However, microscopic images may also feature debris that are usually randomly spread in the image. In addition, pollen grains may be close to each other, forming clusters. In practice, this means that, in both cases, the background of the extracted sub-images is likely to be corrupted with non-pollen grain objects, or other pollen grains (see Figure \ref{fig-grain-seg-intro}). This non-homogeneous background must be discarded by means of mask creation, as it is likely to hamper the quality of the extracted texture features, which may in turn ruin the classification results. Moreover, as mentioned in the introduction, mask creation is a necessary step for the extraction of shape-based features.

\subsection{Coarse stage}

\begin{figure}
 \begin{center}
  $\begin{array}{cccc}
   \multicolumn{1}{l}{} & \multicolumn{1}{l}{} & \multicolumn{1}{l}{} & \multicolumn{1}{l}{}\\
   \includegraphics[scale=0.60]{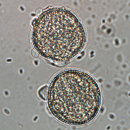} &
   \includegraphics[scale=0.60]{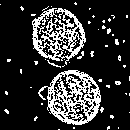} &
   \includegraphics[scale=0.60]{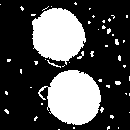} &
   \includegraphics[scale=0.60]{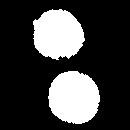}\\
   \mbox{(a)} & \mbox{(b)} & \mbox{(c)} & \mbox{(d)}
  \end{array}$
 \end{center}
\caption{Consecutive steps to automatically segment (a) pollen grains. First, (b) binary classification is used to roughly segment pollen grains. Then, (c) a hole-filling algorithm fills the holes of the inner texture. Finally, (d) opening and closing operations remove the small objects from the image.}
\label{fig-grain-seg-coarse_1}
\end{figure}

First, sub-images of pollen grains are roughly extracted using a procedure involving clustering and morphological operations. As a pre-processing step, a contrast-limited adaptive histogram equalization is applied to enhance the contrast of the image. Then, the image is filtered using a median filter to remove noise while preserving edges. Finally, the coarse segmentation of pollen grains is performed through the following steps (see Figure \ref{fig-grain-seg-coarse_1}).

\begin{enumerate}
  \item Application of a binary classification to the image pixels, so as to roughly separate pollen grains (foreground) from the rest of the image (background). These two classes are determined by the K-Means algorithm and form the binary image $I_b$.
  \item A hole-filling algorithm using 4-connected background neighbors is applied to $I_b$, so as to fill the possible holes featured by the inner texture of pollen grains.
  \item Opening and closing operations are carried out on $I_b$. The goal is to remove small objects from the image, such as debris, while preserving the shape and size of pollen grains.
\end{enumerate}

\subsection{Fine stage}
\label{sec-grain-seg-fine}

After the coarse stage, our experiments have shown that most pollen grains are correctly extracted. However, they have also shown that pollen grains may be either slightly over-segmented (i.e. considering the neighboring background as part of the pollen grain), or slightly under-segmented (i.e. discarding a part of the exine), depending on the exine properties (e.g. size and color). Furthermore, the coarse stage is likely to end up with a non-smooth segmented grain, as the segmentation is only based on a K-Means clustering of intensity values, i.e. with no consideration for the extracted shape. This is why the rough segmentation is followed by a snake-based segmentation \cite{Xu1997}, with external forces attracting the snake to the exine boundary and internal forces ensuring its contour to be smooth.

\begin{figure}
 \begin{center}
  $\begin{array}{ccccc}
   \multicolumn{1}{l}{} & \multicolumn{1}{l}{} & \multicolumn{1}{l}{} & \multicolumn{1}{l}{}\\
   \includegraphics[scale=0.45]{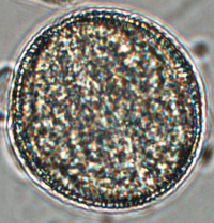} & \includegraphics[scale=0.45]{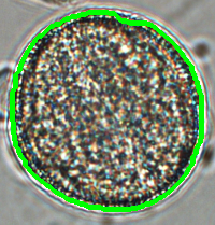} & \includegraphics[scale=0.45]{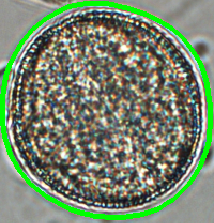} & \includegraphics[scale=0.45]{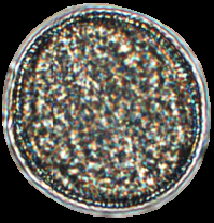}\\[1.5ex]
   \mbox{(a)} & \mbox{(b)} & \mbox{(c)} & \mbox{(d)}\\
  \end{array}$
 \end{center}
\caption{Grain segmentation based on a coarse-to-fine approach: (a) original pollen grain, (b) grain segmentation after coarse stage, (c) grain segmentation after fine stage, and (d) output image after retrieving the background (black mask).}
\label{fig-grain-seg-fine}
\end{figure}

The snake is initialized at the perimeter of the mask generated by the coarse stage. The perimeter is discretized into a set of contour points, which increases the snake's flexibility to fit the exine boundary and speeds up computation time. Empirical tests have shown that keeping one contour point out of 20 consecutive points ensures a good snake initialization. During the segmentation, the snake behaves as a moving contour whose points are attracted to nearby boundaries, such as edges, corners and line terminations. To highlight boundaries, the edge energy image is first extracted by computing the gradient of the original image. Then, a Gradient Vector Flow (GVF), from which the external forces are defined \cite{Xu1997}, is generated from the gradient image. To keep the contour smooth during segmentation, both thin plate energy and balloon force are used as regularization methods. In practice, the snake segmentation consists in 100 iterations, which was found to be sufficient for the snake to fit the grain boundary (see Figure \ref{fig-grain-seg-fine}).

\section{Generalized feature extraction and selection}
\label{sec-features}

\subsection{Feature extraction}
\label{sub-sec-feature-extraction}

\begin{figure}
 \begin{center}
  \includegraphics[scale=0.45]{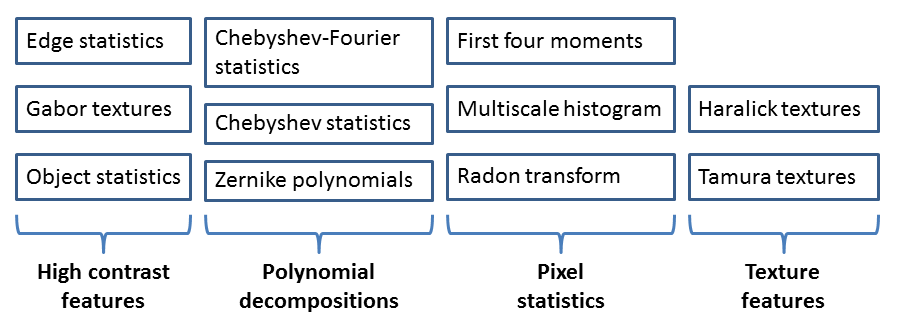}\\
 \end{center}
 \caption{Overview of WND-CHARM features.}
 \label{fig-wndcharm-features}
\end{figure}

Features aim at representing the image content as a set of numeric values. There are mainly two approaches to extract features: \emph{task-specific} and \emph{generalized}. Since our proposed framework aims to be generic, we base the feature extraction on a generalized approach named WND-CHARM \cite{Orlov2008}. The objective is to compute a large number of features and consider them all as potentially discriminant for any given random dataset, rather than selecting a fixed set of features tuned for a specific dataset. The types of features used by WND-CHARM, which are thoroughly described in \cite{Orlov2008}, fall into four categories: high contrast features, polynomial decompositions, pixel statistics and texture features (see Figure \ref{fig-wndcharm-features}).

In addition to calculating these features for the raw image, image pixels are also subject to several standard transforms (i.e. Fourier, wavelet and Chebyshev), from which features are computed. Furthermore, some of these transforms are combined to extract additional features. All features are based either on grayscale or color images. As a result, a feature vector comprising 2164 variables (color images), or 1025 variables (grayscale images), is generated. To complete the intensity-based features proposed by WND-CHARM, 25 additional features based on shape have been added into the feature vector. Morphological features and statistics on shape, such as area, eccentricity, diameter, orientation and perimeter, are computed from the binary image extracted after grain segmentation.

\subsection{Feature selection}
\label{sub-sec-feature-selection}

With generalized feature extraction, the large set of image features provides an extensive numeric description of the image content \cite{Orlov2008}. However, features that are discriminant for one specific dataset may not be discriminant for another dataset, and some features, depending on the dataset, are expected to represent noise. These features are to be considered as irrelevant because they represent useless information likely to degrade the performance of the classification both in terms of speed and accuracy \cite{Guyon2003}.

To select the most discriminative features, WND-CHARM assigns each feature \emph{f} with a Fisher score $W_f$ \cite{Bishop2006}, which is described by the following equation:

\begin{equation}
 \label{fisher_score}
    W_f = \frac{\sum_{c=1}^{N_f} (\overline{T_f} - \overline{T_{f,c}})^2}{\sum_{c=1}^{N_f} \sigma^2_{f,c}} \cdot \frac{N_f}{{N_f}-1}
\end{equation}

where $W_f$ is the Fisher score, $N_f$ is the total number of features, $\overline{T_f}$ is the mean of feature $f$ in the entire dataset, $\overline{T_{f,c}}$ is the mean of feature $f$ in the class $c$, and $\sigma^2_{f,c}$ is the variance of feature $f$ among all samples of class $c$.

All variances are computed after normalization of the features to the interval [0, 100]. Then, they are rank-ordered, so that only features with the highest Fisher scores are taken into account in the classification. Finally, the percentage of the most relevant features must be decided. With microscope pollen images, our empirical tests have shown that using 1-15\% of the strongest features from pollen grain sub-images, i.e. between 10 and 150 features, gives the best results in terms of dataset accuracy.

\section{Pollen grain classification}
\label{sec-classification}

For the classification, or \emph{recognition}, of an unknown pollen grain whose type needs to be identified as one of the pollen types from the training dataset, classifiers such as the Minimum Distance Classifier (MDC) \cite{Rodriguez2004}, Support Vector Machine (SVM) \cite{Rodriguez2006} or Neural Networks (NNs) \cite{France2000} have been considered in the literature and successfully implemented for a variety of applications related to pollen recognition. In practice, features found to be discriminative when building the training dataset are then extracted from the unknown grain and compared with those from the training dataset. The classification stage consists in attributing the unknown grain to the most similar pollen type. In this work, we implemented and compared four classifiers: Weighted Neighbor Distance (WND-5), which is a variation of the Nearest Neighbor classifier, feed-forward back-propagation Neural Network (NN), Decision Tree (DT) and Random Forest (RF).

\subsection{Weighted Neighbor Distance}

WND-CHARM classifies the feature vectors of given test samples using a variation of the Nearest Neighbor classifier called Weighted Neighbor Distance (WND-5) \cite{Orlov2008}. For feature vector $z$ computed from a test image, the distance $d_{z,c}$ of the image from a given class $c$ is measured by the following equation:

\begin{equation}
 \label{fisher_score}
    d_{z,c} = \frac{\sum_{t \in T_c} \left(\sum_{f=1}^{|z|} W_f (z_f - t_f)^2 \right)^p}{|T_c|}
\end{equation}

where $T_c$ is the training set of class $c$, $t$ is a feature vector from $T_c$, $|z|$ is the length of feature vector $z$, $z_f$ is the value of image feature $f$ in the vector $z$, $W_f$ is the Fisher score of feature $f$, $|T_c|$ is the number of training samples of class $c$, and $p$ is the exponent set to -5 (this value has been determined empirically \cite{Orlov2008}).

The distance between a feature vector and a given class is the mean of its weighted distances (to the power of $p$) to all feature vectors of that class. After computing the distances from sample $z$ to all classes, the class that has the shortest distance is the classification result. While in classical Nearest Neighbor \cite{Duda2001} only the closest (or $k$ closest) training samples determine the class of a given sample, WND-5 measures the weighted distances from the given sample to all training samples of each class, so that all samples in the training set can affect the classification result. This modification of the traditional Nearest Neighbor has been reported to provide a more accurate classification \cite{Orlov2008}.

\subsection{Decision Tree and Random Forest}

A Decision Tree (DT) is a conceptually simple, yet robust and widely used tool for decision support in which the classification is performed through a tree graph \cite{Breiman2001}. The classification starts from an initialization node (\emph{root node}) from which a given test sample is tested at each stage (\emph{internal node}) of the classification, all the way down to the end of a tree branch (\emph{leave} or \emph{terminal node}) \cite{Tan2006}. The path followed by the sample depends on threshold-based conditions associated to each internal node.

To select the optimal threshold-based conditions, DT algorithms make use of a brute force method, which consists in testing all potential variables and selecting the variable that maximizes a given criterion. When building the DT, this criterion characterizes the quality of the split created by the transition from an internal node to its associated leaves. There are a large number of criteria based on information theory or statistics, such as the Shannon entropy and the Gini coefficient \cite{Tan2006}.

As for the classification in general, the final aim of the DT is to represent at best the training dataset while ensuring an optimal generalization of the data. When the structure is too complex, with lots of branches and internal nodes, the training dataset is \emph{too well} represented and the DT unlikely to generalize new data, which is the final objective of any classifier. In this case, we talk about \emph{over-fitting}. Conversely, when the DT is too simple, new data are likely to be better represented, but at a cost of a poorer segmentation performance. In this case, we talk about \emph{under-fitting}. Therefore, the overall objective is to find an optimal trade-off between over-fitting and under-fitting. To do so, the idea is to build a DT as small as possible while ensuring an optimal segmentation performance.

In the literature, the common method to optimize this trade-off consists in creating testing datasets and using them to test the performance of the DT. The optimal complexity may be found by performing a parameter optimization using DTs of increasing complexity, DTs built from training datasets of increasing size, or different testing datasets. Most complex methods include the Vapnik-Chervonenkis (VC) criterion \cite{Vapnik1971}, which searches for the optimum between training and testing dataset error. However, the most common methods are based on pruning, either during (\emph{pre-pruning}) or after (\emph{post-pruning}) DT construction. In the former case, the method consists in using stopping criteria to stop the DT construction before it reaches over-fitting. In the latter case, the construction is done in two stages. First, the DT is built to be as accurate as possible from a subset of the training dataset called the \emph{growing set}. Then, the classification performance is improved by pruning the DT using the other part of the dataset called the \emph{pruning set}.

To improve classification accuracy and robustness, the Random Forest (RF) classifier, which is built upon an ensemble of DTs, has been proposed in the literature \cite{Breiman2001}. During the training stage, the DTs learn from different subsets of the training dataset and no pruning is performed after their construction. Each DT is built using the values of random feature vectors in a way that all DTs from the RF possess the same distribution. The random feature vectors may be generated using several techniques, such as bagging \cite{Breiman2001}, random split selection \cite{Dietterich2000} and random subspace method \cite{Ho1998}. When classifying an unknown sample, its feature vector is tested using all DTs of the RF. Their outputs constitute votes for the most popular class, which in turn is the RF prediction. Nowadays, RF classifier is considered as the most accurate learning algorithm and its performance has been proven on many datasets \cite{Caruana2008}.

\subsection{Neural Network}

A neural network (NN) is a computational model whose design is very schematically inspired from the operations of brain's biological neurons. As said in the introduction, NNs have been widely used for pollen classification. Their first implementation for pollen classification consisted in a feed-forward NN with Haralick texture features as input data \cite{Li1999}. More sophisticated NNs were then presented in the literature. The Pattern Recognition Architecture for Deformation Invariant Shape Encoding (Paradise) \cite{France2000}, which had been initially designed to recognize visual objects, such as hand gestures, faces and handwritten numerals, comprises three layers: one for feature extraction, one for pattern detection and one for classification. The Multi-Layer Perceptron (MLP), which is a feed-forward NN trained by the back-propagation algorithm, was also presented \cite{Zhang2004}, although with only one single layer used. More recently, MLP performance was compared with both Minimum Distance Classifier (MDC) and Support Vector Machine (SVM) \cite{Rodriguez2006}.

As for DTs, the objective of NNs, when used for supervised learning, consists not only in building a network that is able to represent at best training data, but also in being able to generalize new data. Common methods to train the NN, such as the mean-squared error and the gradient descent, make use of an optimization based on a cost function, whose role is to quantify the NN's ability to represent the training dataset. A typical approach to avoid over-fitting consists in a cross-validation scheme in which a testing dataset is used to optimize the NN parameters such as to minimize the generalization error.

\section{Authentication}
\label{sec-authentication}

In the literature, the authentication problem is also referred to as \emph{outlier detection} \cite{Hodge2004,Aggarwal2001}, \emph{novelty detection} \cite{Markou2003a,Markou2003b}, and \emph{concept learning} \cite{Goodman2008,Hekanaho1997}. In our case, the authentication of pollen types is a much more complex task than their classification, although both tasks are based on the same methods, i.e. features are extracted, then selected (if necessary), and finally classified.

\begin{figure}
 \begin{center}
  \includegraphics[scale=0.30]{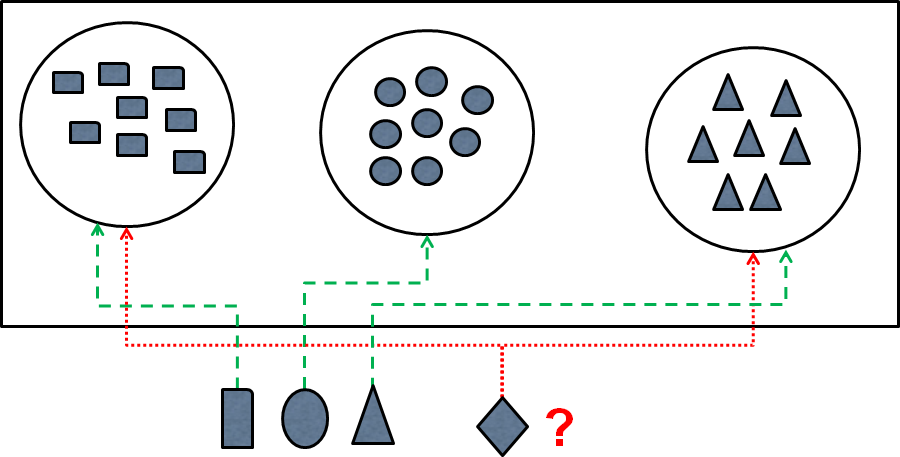}\\
 \end{center}
 \caption{To illustrate the difference between classification and authentication, note how deformed square, circle and triangle shapes have no difficulties to be associated to their respective class (classification, green dashed lines). However, the diamond shape, which is not represented by any class, is likely to be misclassified, i.e. associated either to the square or the triangle class (authentication, red dotted lines), and therefore needs to be identified as belonging to an unknown class.}
 \label{fig-authentication-scheme}
\end{figure}

In a classification scheme, all pollen types are known (which is a requirement for the construction of the training dataset). When an unknown pollen grain needs to be identified, the classifier performs the classification by associating it to the most similar pollen type from the training dataset. In this case, the classifier makes the assumption that the pollen grain belongs to one pollen type from the training dataset. However, in an authentication scheme, the unknown pollen grain may belong to an unknown pollen type. In this case, the classifier needs not only to find out which pollen type the unknown pollen grain belongs to (should it belong to one of them), but also needs to find out whether the pollen grain belongs to a pollen type from to the training dataset (see an illustrative example in Figure \ref{fig-authentication-scheme}).

As explained in Section \ref{subsec-classification}, the RF classifier gives the best classification results, which is why this classifier has been chosen for the pollen authentication. During classification, each DT votes for a class (i.e. pollen type). Our pollen authentication system consists in defining a dynamic threshold on the number of votes received by each class, from which an unknown pollen grain is either considered as \emph{inlier} (i.e. belonging to one of the training dataset's pollen types), or \emph{outlier} (i.e. belonging to a pollen type different from the training dataset's pollen types). The threshold is dynamic because its value is adaptive, depending on the kind, but also on the number, of pollen types constituting the training dataset.

Four conditions on the classification results, denoted as \textit{$\theta_{11}$}, \textit{$\theta_{12}$}, \textit{$\theta_{21}$}, and \textit{$\theta_{22}$}, and which are defined in Equation \ref{eq_theta-11}, Equation \ref{eq_theta-12}, Equation \ref{eq_theta-21}, and Equation \ref{eq_theta-22}, respectively, have been studied for the authentication.

 \begin{equation}
  \theta_{11} = Vp_1(x) > \mbox{MIN}(\mbox{TP}(p_1))
 \label{eq_theta-11}
 \end{equation}

where $p_1$ is the pollen type that has received the highest number of votes $Vp_1(x)$ for the classification of unknown pollen grain $x$, and $\mbox{TP}(p_1)$ is the set of votes received by $p_1$ during testing stage for the correct classification of its associated pollen grains from the testing dataset (true positives).

 \begin{equation}
  \theta_{12} = \theta_{11} \mbox{ AND } ((Vp_1(x) - Vp_2(x)) > (\mbox{MIN}(\mbox{TP}(p_1)-\mbox{TP}(p_2))))
 \label{eq_theta-12}
 \end{equation}

where $p_1$ and $p_2$ are, respectively, the pollen types that have received the first and second highest number of votes $Vp_1(x)$ and $Vp_2(x)$ for the classification of unknown pollen grain $x$, and $\mbox{TP}(p_1)$ and $\mbox{TP}(p_2)$ are, respectively, the set of votes received by $p_1$ and $p_2$ during testing stage for the correct classification of their associated pollen grains from the testing dataset (true positives).

 \begin{equation}
  \theta_{21} = Vp_1(x) > (\mbox{MEAN}(\mbox{TP}(p_1)) - \mbox{STD}(\mbox{TP}(p_1)))
 \label{eq_theta-21}
 \end{equation}

where $p_1$ is the pollen type that has received the highest number of votes $Vp_1(x)$ for the classification of unknown pollen grain $x$, $\mbox{TP}(p_1)$ is the set of votes received by $p_1$ during testing stage for the correct classification of its associated pollen grains from the testing dataset (true positives), and STD is the standard deviation.

 \begin{equation}
  \theta_{22} = \theta_{21} \mbox{ AND } ((Vp_1(x) - Vp_2(x)) > (\mbox{MIN}(\mbox{TP}(p_1)-\mbox{TP}(p_2))))
 \label{eq_theta-22}
 \end{equation}

where $p_1$ and $p_2$ are, respectively, the pollen types that have received the first and second highest number of votes $Vp_1(x)$ and $Vp_2(x)$ for the classification of unknown pollen grain $x$, and $\mbox{TP}(p_1)$ and $\mbox{TP}(p_2)$ are, respectively, the set of votes received by $p_1$ and $p_2$ during testing stage for the correct classification of their associated pollen grains from the testing dataset (true positives).

\section{Results}

\subsection{Classification}
\label{subsec-classification}

We have tested the classification accuracy of our general framework on multi-focal microscope images acquired with the material and procedures described in Section \ref{sec-material}. These images were gathered in a microscope image database composed of the 15 pollen types presented in Section \ref{sec-data}. For each microscope image, 31 consecutive focal planes have been acquired and the optimal focal image has been selected using the absolute gradient method \cite{Redondo2011}, which has proven in our experiments to be both time-efficient and visually accurate compared to the other methods presented in Section \ref{sec-foc-selection}. Then, pollen grain sub-images have been extracted from the optimal focal images using the automatic grain segmentation presented in Section \ref{sec-grain-seg}. Manual grain segmentation has been performed when automatic segmentation failed. Each pollen type of the database $S$ is considered as a different class and gathers 120 microscope sub-images for a total of 1800 pollen grain sub-images.

To test the robustness of our framework, we have created sub-datasets $S^\prime_p$ of increasing size $p$ ranging from 2 to 15 pollen types (i.e. $p$ = \{2, 3, 4, \ldots, 15\}) for a total of 14 consecutive dataset sizes. For each $S^\prime_p$, pollen types are selected in a random fashion (except for $p = 15$, in which case all pollen types from $S$ are selected). To test the reproducibility of our framework, classification accuracy has been calculated as the mean $\pm$ SD computed on 10 different ${S^\prime_p}$. Finally, for each $S^\prime_p$, the 120 microscope sub-images associated to each pollen type have been randomly separated using a Leave-One-Out (LOO) approach, i.e. one training dataset (to train the sub-dataset) and one testing dataset (to calculate the classification accuracy) using a $\frac{3}{4}/\frac{1}{4}$ ratio, i.e. 90 sub-images for training and 30 sub-images for testing.

For each sub-dataset $S^\prime_p$, both texture and shape features have been extracted as depicted in Section \ref{sub-sec-feature-extraction}. Features from the training dataset have been trained for Neural Network (NN), Decision Tree (DT) and Random Forest (RF) classifiers. There is no training stage associated with the Weighted Neighbor Distance (WND-5) classifier, since the classification consists in minimizing the distance between the feature vector of an unknown pollen grain and all feature vectors from the training dataset. For NN, the Scaled Conjugate Gradient algorithm \cite{Moller1993} has been used as training function. No hidden layers have been implemented as their number, and the number of their associated neurons, require a specific optimization that depends on the input data. Such an optimization, which has been so far tuned manually for pollen classification \cite{Zhang2004,France2000,Li1999}, would not comply with the generic purpose of our framework. Therefore, only two layers have been implemented: the input layer with the number of neurons corresponding to the number of features, and the output layer with as many neurons as pollen types (i.e. classes). DTs have been constructed using both the Gini's diversity index as a split criterion and a post-pruning step \cite{Breiman2001}. As for RFs, they have been trained using the same parameters and 500 DTs \cite{Breiman2001}. For both WND-5 and NN classifiers, the most relevant and discriminant features have been selected using the Fisher score presented in Section \ref{sub-sec-feature-selection}. As for DT and RF classifiers, there is no need for feature selection prior to the training stage since all features are needed during DT construction.

\begin{figure}
 \begin{center}
  $\begin{array}{ccc}
   \multicolumn{1}{l}{} & \multicolumn{1}{l}{}\\
   \includegraphics[scale=0.45]{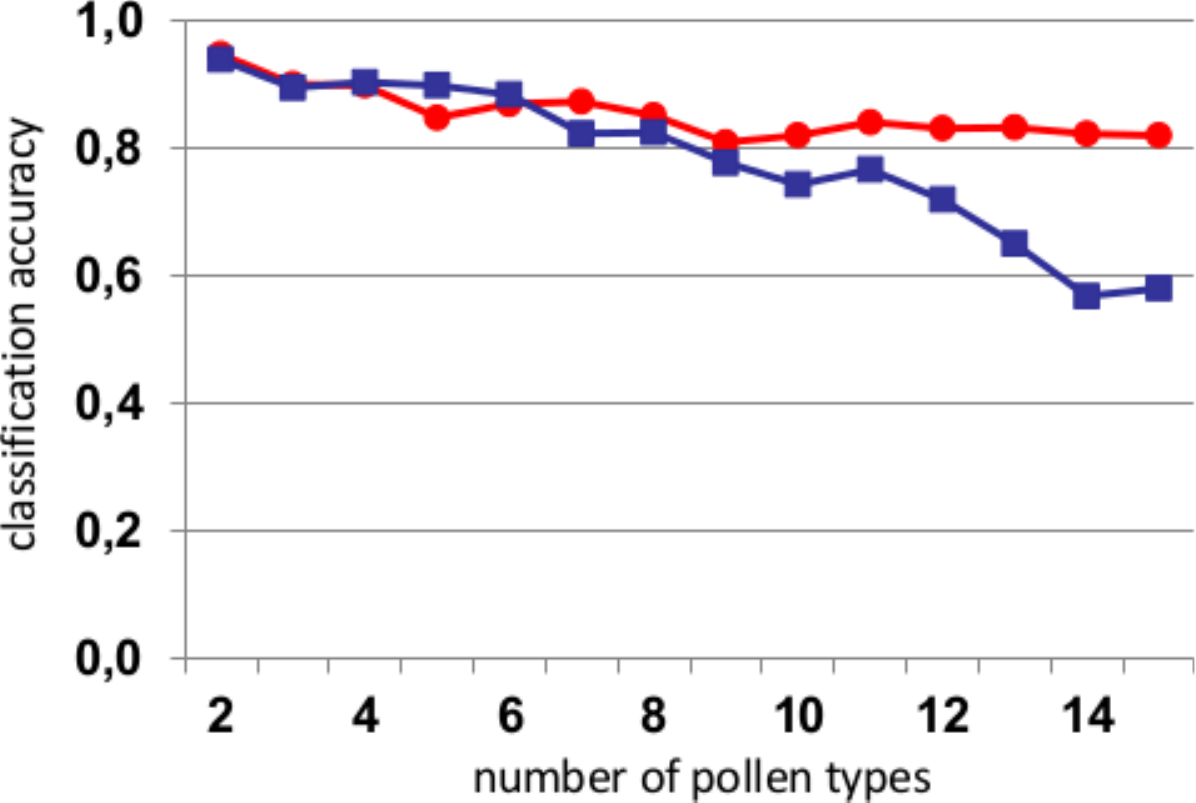} & \includegraphics[scale=0.45]{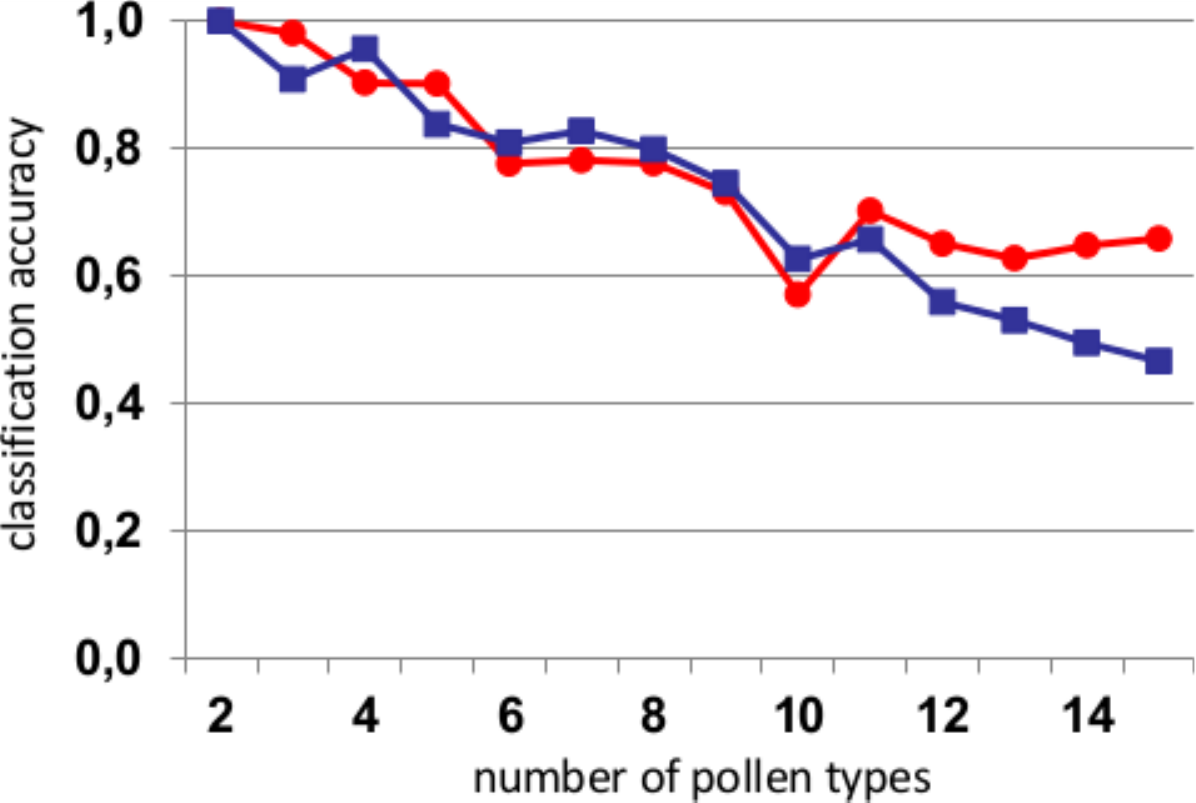}\\
   \mbox{(a)} & \mbox{(b)}\\
   \includegraphics[scale=0.45]{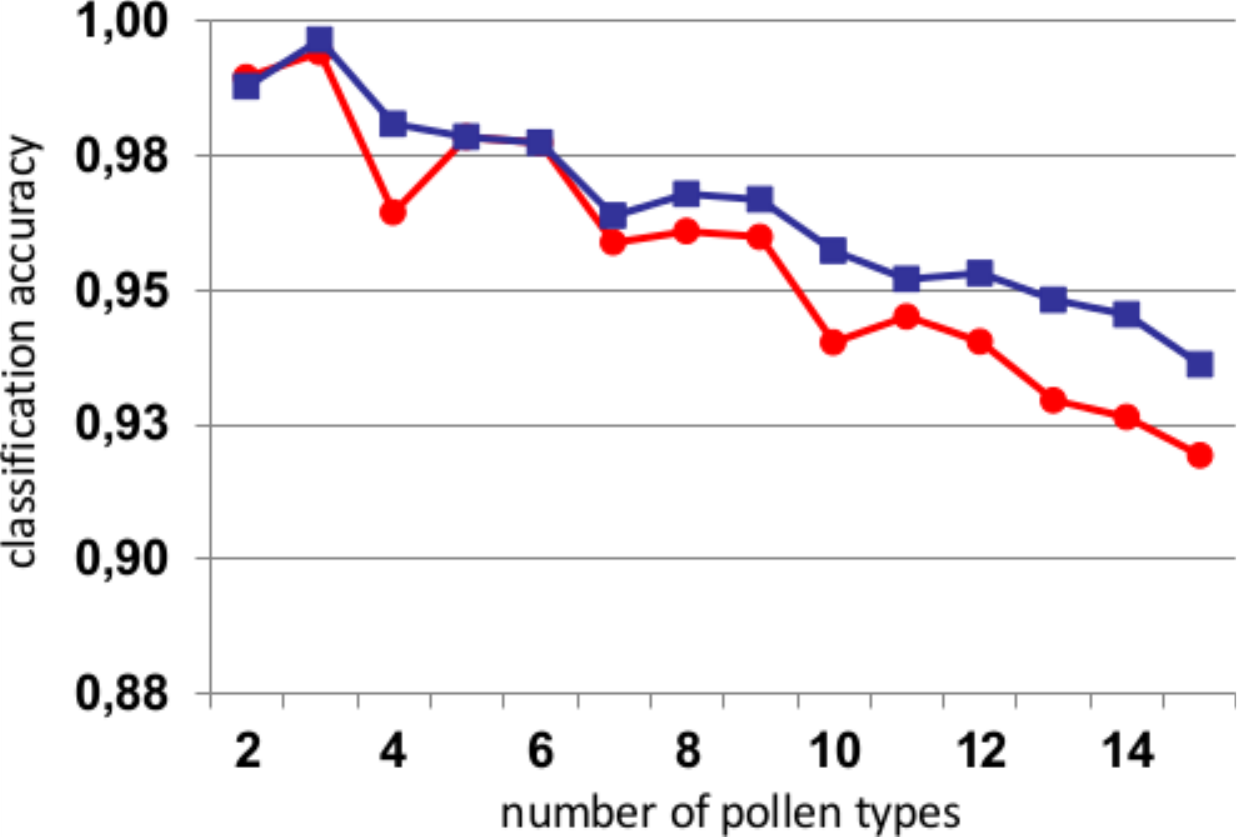} & \includegraphics[scale=0.45]{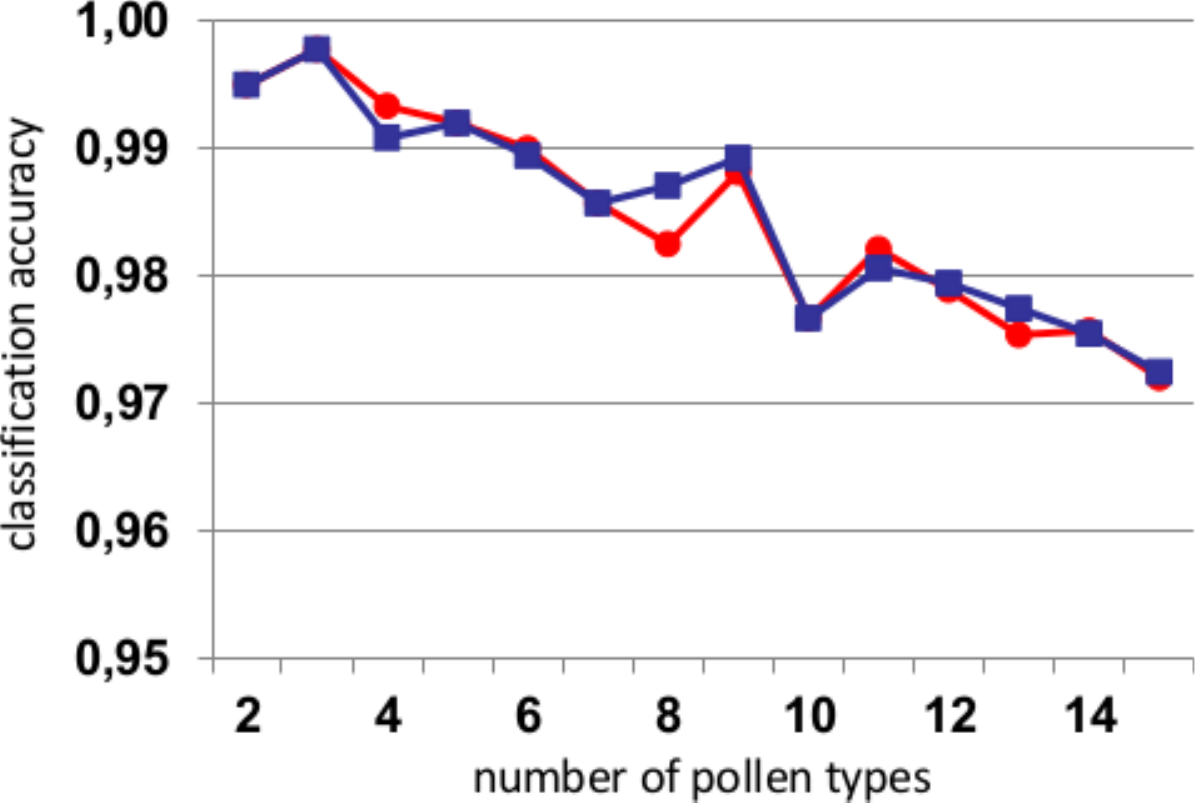}\\
   \mbox{(c)} & \mbox{(d)}\\
  \end{array}$
 \end{center}
\caption{Classification accuracy $\alpha$ (ordinate) with respect to the number of pollen types $p$ included in the training dataset (abscissa) when using: (a) WND-5, (b) NN, (c) DT, and (d) RF classifiers. Red circle line depicts $\alpha$ when using only intensity-based features ($\alpha_i$), and blue square line depicts $\alpha$ when using them with shape-based features ($\alpha_{i+s}$). Both $\alpha_i$ and $\alpha_{i+s}$ have been calculated as the mean computed on 10 different sub-datasets $S^\prime_p$.}
\label{fig-texture-shape}
\end{figure}

We have compared the classification accuracy $\alpha$ of the four classifiers with respect to the number of pollen types $p$ included in the training dataset. Moreover, we have studied the influence of the small number of shape-based features (i.e. 25) with respect to the large number of intensity-based features (i.e. 1025) by comparing $\alpha$ when using only intensity-based features ($\alpha_i$), and when using them with shape-based features ($\alpha_{i+s}$). Comparative results are depicted in Figure \ref{fig-texture-shape}.

For all classifiers, $\alpha$ decreases when $p$ increases, which is an expected result as the more pollen types in ${S^\prime_p}$, the higher the probability for an unknown type from the testing dataset to get confused during classification. However, this decrease is not the same for all classifiers. For both WND-5 and NN, $\alpha_i$ and $\alpha_{i+s}$ undergo a quite similar decrease until $p = 10$, from which $\alpha_i$ seems to get more stable while $\alpha_{i+s}$ keeps decreasing. Nonetheless, WND-5 is more robust when increasing $p$, as we have $\alpha_i = 0.82$ (WND-5) and $\alpha_i = 0.66$ (NN) with $p = 15$. As for both DT and RF, results are clearly better, as we have $\alpha_{i+s} = 0.97$ (DT) and $\alpha_{i+s} = 0.98$ (RF) with $p = 15$. Although the influence of shape-based features is not significant for RF, they slightly increase accuracy for DT, especially from $p = 7$. Overall, RF presents the best classification results with $\alpha > 0.97$, regardless of $p$.

\begin{figure}
 \begin{center}
  $\begin{array}{ccc}
   \multicolumn{1}{l}{} & \multicolumn{1}{l}{}\\
   \includegraphics[scale=0.45]{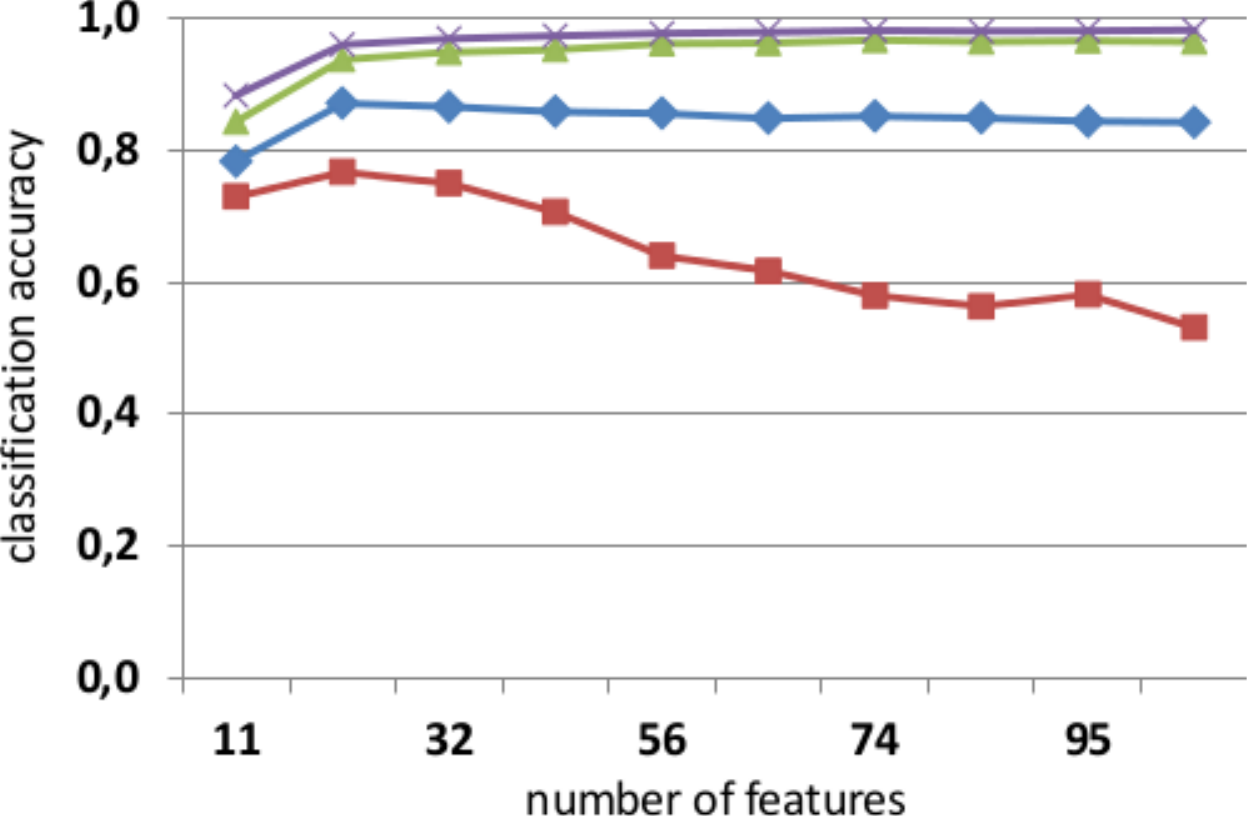} & \includegraphics[scale=0.45]{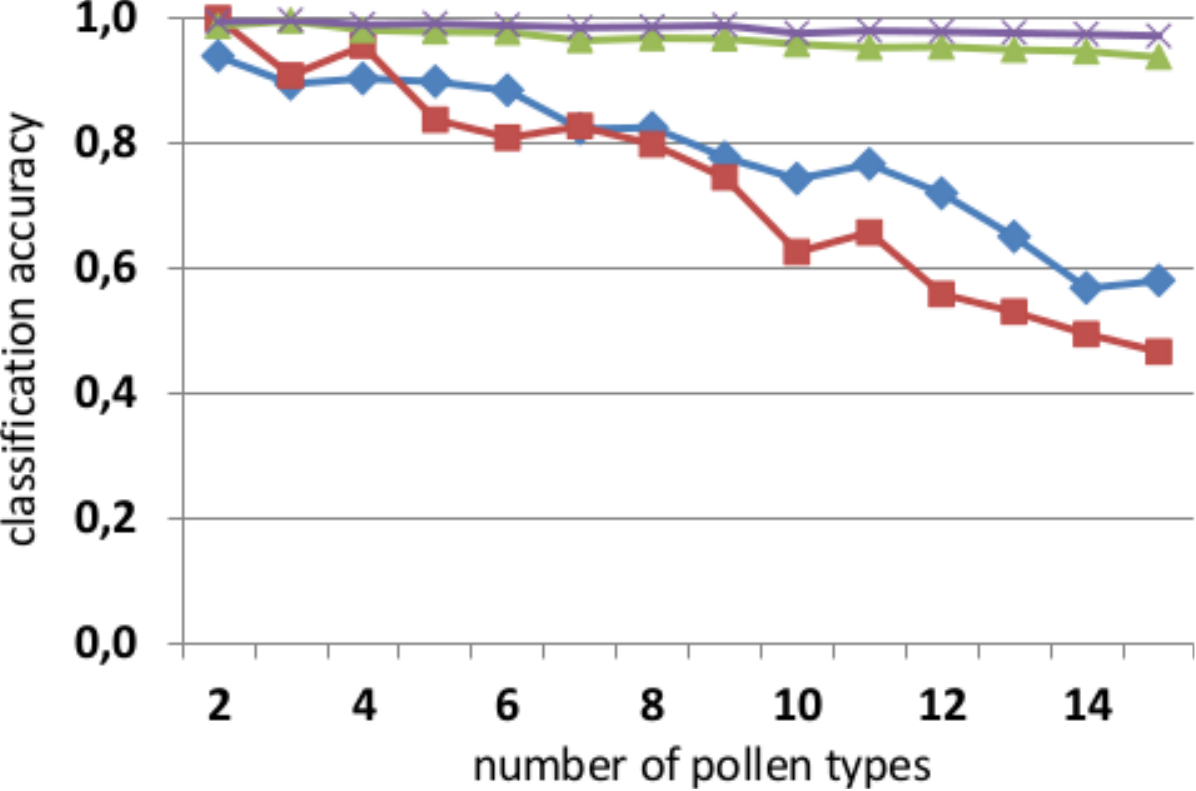}\\
   \mbox{(a)} & \mbox{(b)}\\
  \end{array}$
 \end{center}
\caption{Classification accuracy $\alpha$ (ordinate) with respect to: (a) the number of features $n_f$ (abscissa), and (b) the number of pollen types $p$ included in the training dataset (abscissa) when using: WND-5 (blue diamond), NN (red square), DT (green triangle), and RF (purple cross) classifiers. In (a), $\alpha$ has been calculated as the mean computed on 10 different sub-datasets ${S^\prime_p}$, each of them with $p$ = \{2, 3, 4, \ldots, 15\} for a total of 14 combinations. In (b), $\alpha$ has been calculated as the mean computed on 10 different sub-datasets ${S^\prime_p}$ and with number of features $n_f = 20$ for both WND-5 and NN classifiers.}
\label{fig-comp-features}
\end{figure}

To study the influence of the number of features $n_f$ on $\alpha$, we have launched the four classifiers on $S^\prime_p$ with an increasing number of features, from 1 to 10\% of $N_f$ = 1050, i.e. $n_f$ = 10 to 105 features (see Figure \ref{fig-comp-features}, left). For both WND-5 and NN, the higher $n_f$, the worst $\alpha$, except for $n_f = 20$, which appears to be the optimal number of features. Conversely, for both DT and RF, the higher $n_f$, the better $\alpha$. This is because the feature selection is performed during training stage (i.e. DT construction), and a larger set of features means a higher probability to find the optimal threshold-based condition associated to each DT node. Using the optimal $n_f = 20$ for both WND-5 and NN, and $n_f = N_f = 1050$ (all features) for both DT and RF, we have launched a final classification with increasing number of pollen types $p$ (see Figure \ref{fig-comp-features}, right). Once again, both DT and RF feature a far more robust behavior with respect to $p$, although with RF getting slightly better than DT as $p$ increases.


\subsection{Authentication}
\label{subsec-authentication}

In addition to the training dataset, comprising the same 15 pollen types and 1800 images used for the classification tests (see Section \ref{sec-data}), we created two additional datasets to test our authentication method: an \textit{inlier} and \textit{outlier} dataset (see Table \ref{tab-datasets} for details about the number of images included in the training, inlier, and outlier dataset). The inlier dataset comprises 280 pollen images whose 7 types (i.e. \textit{Brassica}, \textit{Castanea}, \textit{Cistus}, \textit{Echium}, \textit{Olea}, \textit{Quercus}, and \textit{Salix}) belong to the training dataset, and the objective is to test if the RF-based classifier is able to authenticate them as known pollen types, or \textit{inliers}. Conversely, the outlier dataset comprises 280 pollen images whose 7 pollen types (i.e. \textit{Anthemis}, \textit{Apiaceae}, \textit{Citrus}, \textit{Citrus Asia}, \textit{Hedera}, \textit{Papaver}, and \textit{Platanus}) do not belong to the training dataset, and, in this case, the objective is to test if the RF-based classifier is able to authenticate them as unknown pollen types, or \textit{outliers}.

\begin{table}[t]
\centering
\begin{tabular}{lccc}
Pollen type & Training dataset & Inlier dataset & Outlier dataset\\
\hline
\hline
Anthemis	& - & - & 40\\
\hline
Apiaceae		& - & - & 40\\
\hline
Aster	& 120 & - & -\\
\hline
Brassica	& 120 & 40 & -\\
\hline
Campanulaceae		& 120 & - & -\\
\hline
Carduus			& 120 & - & -\\
\hline
Castanea			& 120 & 40 & -\\
\hline
Cistus		& 120 & 40 & -\\
\hline
Citrus		& - & - & 40\\
\hline
Citrus Asia		& - & - & 40\\
\hline
Cytisus			& 120 & - & -\\
\hline
Echium			& 120 & 40 & -\\
\hline
Ericaceae	& 120 & - & -\\
\hline
Hedera	& - & - & 40\\
\hline
Helianthus	& 120 & - & -\\
\hline
Olea	& 120 & 40 & -\\
\hline
Papaver	& - & - & 40\\
\hline
Platanus	& - & - & 40\\
\hline
Prunus	& 120 & - & -\\
\hline
Quercus	& 120 & 40 & -\\
\hline
Salix	& 120 & 40 & -\\
\hline
Teucrium	& 120 & - & -\\
\hline
\hline
TOTAL	& 1800 & 280 & 280\\
\end{tabular}
\caption{For each pollen type, number of pollen grain sub-images included in the training, inlier, and outlier dataset. Both inlier and outlier datasets are used to test our authentication method.}
\label{tab-datasets}
\end{table}

\begin{figure}
 \begin{center}
  $\begin{array}{ccc}
   \multicolumn{1}{l}{} & \multicolumn{1}{l}{}\\
   \includegraphics[scale=0.45]{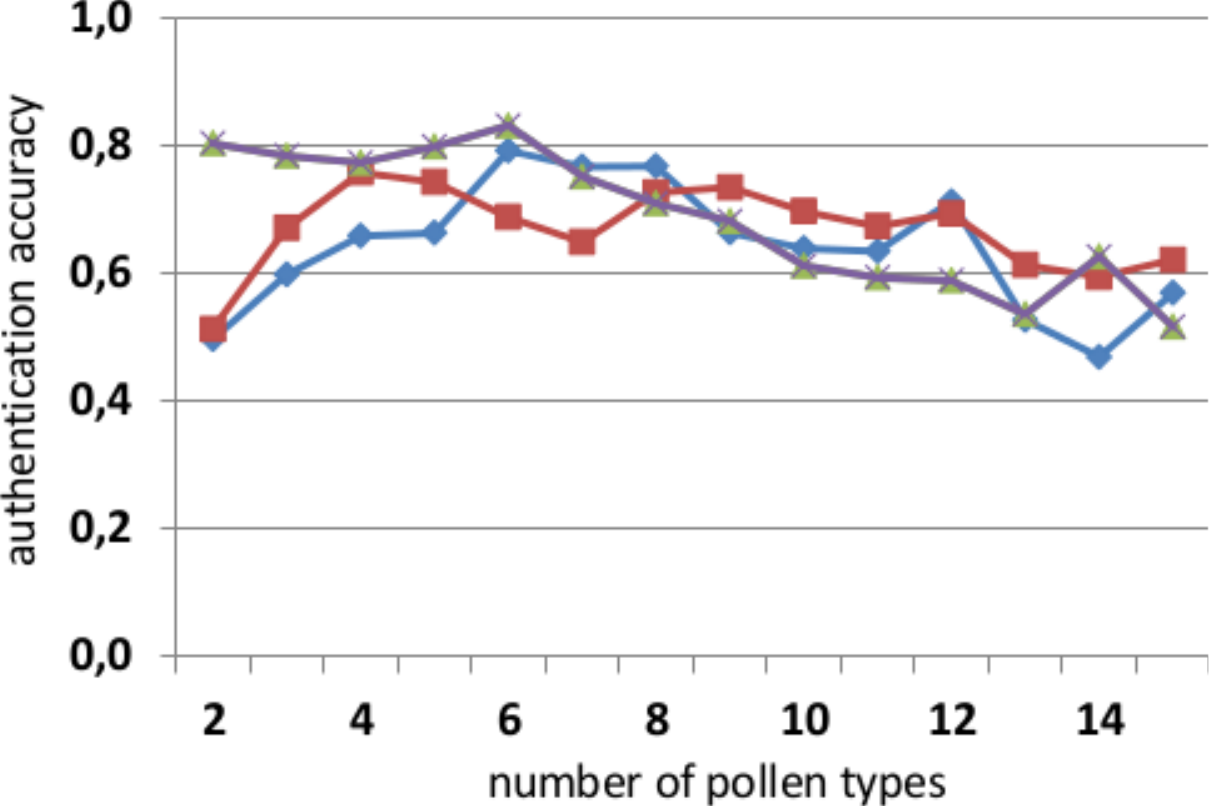} & \includegraphics[scale=0.45]{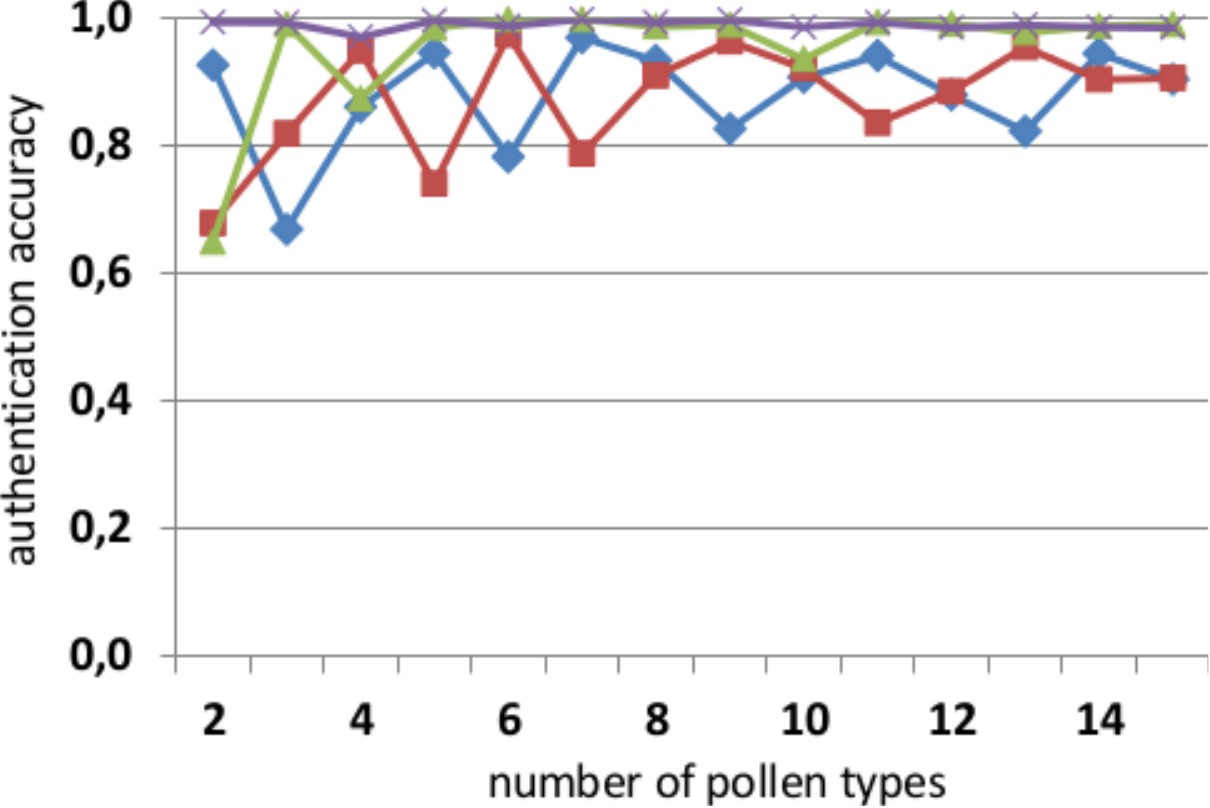}\\
   \mbox{(a)} & \mbox{(b)}\\
  \end{array}$
 \end{center}
\caption{(a) Inlier and (b) outlier authentication accuracy (ordinate) with respect to the number of pollen types $p$ included in the training dataset (abscissa) when using: $\theta_{11}$ (red square), $\theta_{12}$ (blue diamond), $\theta_{21}$ (purple cross), and $\theta_{22}$ (green triangle) conditions (see Section \ref{sec-authentication}). Authentication accuracy has been calculated as the mean computed on 10 different sub-datasets $S^\prime_p$.}
\label{fig-authentication}
\end{figure}

As for the classification tests (see Section \ref{subsec-classification}), we tested the robustness of our authentication framework by creating sub-training datasets $S^\prime_p$ of increasing size $p$ ranging from 2 to 15 pollen types (i.e. $p$ = \{2, 3, 4, $\ldots$, 15\}) for a total of 14 consecutive training dataset sizes. For each $S^\prime_p$, pollen types are selected in a random fashion (except for $p$ = 15, in which case all pollen types from $S$ are selected). To test the reproducibility of our framework, authentication accuracy has been calculated as the mean $\pm$ SD computed on 10 different $S^\prime_p$. For each sub-dataset $S^\prime_p$, only intensity-based features have been extracted, since the influence of shape-based features has shown not to be significant for the RF classifier (see Section \ref{subsec-classification}). Comparative results for both inlier and outlier authentication accuracy are depicted in Figure \ref{fig-authentication}a and Figure \ref{fig-authentication}b, respectively.

Regarding the inlier authentication accuracy $\alpha_{\mbox{in}}$ (Figure \ref{fig-authentication}a), the four conditions on the classification results ($\theta_{11}$, $\theta_{12}$, $\theta_{21}$, and $\theta_{22}$) seem to depict a quite similar behavior. However, with $\mu(\alpha_{\mbox{in}})$ being the mean of $\alpha_{\mbox{in}}$ calculated on all number of pollen types $p$, $\theta_{21}$ and $\theta_{22}$ (both $\mu(\alpha_{\mbox{in}})$ = 0.68) feature a slightly better accuracy with respect to $\theta_{11}$ ($\mu(\alpha_{\mbox{in}})$ = 0.67) and $\theta_{12}$ ($\mu(\alpha_{\mbox{in}})$ = 0.64). Regarding the outlier authentication $\alpha_{\mbox{out}}$ (Figure \ref{fig-authentication}b), although the four conditions on the classification results show better results compared to the inlier authentication, $\theta_{21}$ ($\mu(\alpha_{\mbox{out}})$ = 0.99) clearly features the best accuracy with respect to the others, i.e. $\theta_{11}$ ($\mu(\alpha_{\mbox{out}})$ = 0.87), $\theta_{12}$ ($\mu(\alpha_{\mbox{out}})$ = 0.88) and $\theta_{22}$ ($\mu(\alpha_{\mbox{out}})$ = 0.95).

Overall, these results show that we have two different scenarios, one optimizing the inlier authentication (i.e. using either $\theta_{21}$ or $\theta_{22}$ condition) and the other one optimizing the outlier authentication (i.e. using $\theta_{21}$ condition). However, in an outlier detection scheme, one usually prefers to be conservative and make sure that no outliers are considered as inliers (even at the cost of discarding inliers that could be considered as outliers), rather than optimizing the number of identified inliers (which, in this case, is at the risk of considering outliers as inliers). This is why we have decided to base the authentication part of our framework on the $\theta_{21}$ condition.

\section{Conclusion}

To our knowledge, this work is the first study of the classification accuracy when combining both a generalized feature extraction and an increasing number of pollen types, or \emph{taxa}. We have presented a brute force-like general framework for the classification of multi-focal images and applied the method on microscope pollen images. Unlike previous pollen classification methods optimized for, and making use of, a fixed number of pollen types (e.g. 3 \cite{Rodriguez2006,France2000}, 5 \cite{Zhang2004}, 8 \cite{Ranzato2007} and 9 \cite{Mitsumoto2009}), our framework has been designed to be efficient regardless of the number of pollen types. So far, in the literature, focus has been put on the classification accuracy and the majority of recent publications have presented very good results. However, we believe that this accuracy, although still an important parameter in pollen classification, should be coupled with the system robustness to deal with any kind, and any number, of pollen types, so as to be successfully implemented in daily routine.

From the four classifiers presented in this work, Random Forest (RF) clearly features the best and most robust classification results with an accuracy above 97\% for any number of pollen types. These are very encouraging results for a practical use of pollen classification, which would require both accuracy and robustness. These two fundamental properties featured by RFs are explained by their inherent generic construction. First, they do not require a feature selection prior to the training stage as all features are considered during tree construction. This property is optimally coupled with generalized feature extraction computing a very large number of features. Second, the pruning method associated with RF ensures the efficient removal of non-discriminant features for some Decision Trees (DTs), while reconsidering them for other DTs of the RF, where they could be possibly discriminant. Regarding both Weighted Neighbor Distance (WND-5) and Neural Network (NN) classifiers, accuracy is significantly degraded by the number of pollen types. For WND-5, this is explained by the lack of training stage associated to this Nearest Neighbor-based classifier, which makes it less robust when using a large number of pollen types. For NNs, there are two main reasons. First, their basic architecture with no hidden layers, while complying with the generic purpose of our framework, does not create a network robust enough to deal with a large number of pollen types. Second, their inherent application-specific construction makes them unsuitable for a generic approach. Unlike DT and RF, NN accuracy strongly depends on the selected features prior to the training stage. Also, their construction (e.g. number of hidden layers and neurons) must usually be optimized for a specific application (e.g. recognition of a set of known pollen types), as it is demonstrated in the most significant works relying on NN \cite{Zhang2004,France2000,Li1999}. We believe that the good performance of NN published so far is mainly due either to the relative small number of pollen types, or to the choosing of both discriminant features and optimized network configuration.

Regarding pollen authentication, we have extended our general pollen classification framework in a way that it can detect if unknown pollen types are either \emph{known} (classification), or \emph{unknown} (authentication), to the training dataset. To do so, we have combined the RF classifier, which has proven to give the best and most robust classification results, with four conditions on the classification results. In practice, after classification, the votes of each DT are used to define a classification boundary around the positive class, maximizing the number of \emph{inliers} (known pollen types) while rejecting \emph{outliers} (unknown pollen types). Results from Section \ref{subsec-authentication} show that $\theta_{21}$ condition features the best accuracy for the authentication of both inliers and outliers. Although our authentication framework based on $\theta_{21}$ gives excellent results for the outlier authentication ($\mu(\alpha_{\mbox{out}})$ = 0.99), its restrictive threshold causes the rejection of a non-negligible number of inliers ($\mu(\alpha_{\mbox{in}})$ = 0.68). However, in an outlier detection scheme, one usually prefers to be conservative and make sure that no outliers are considered as inliers (even at the cost of discarding inliers that could be considered as outliers), rather than optimizing the number of identified inliers (which, in this case, is at the risk of considering outliers as inliers).

Overall, the objective of this article is to propose a general framework for multi-focal image classification and authentication. Although we have implemented a brute force-like approach designed to be efficient on any kind of pollen images, its generic design ensures it to be used on any kind of multi-focal images, such as biological cells [29]. Indeed, all stages of the framework's pipeline have been implemented in an automatic and generic fashion: the optimal focus selection stage using the absolute gradient method, the segmentation stage involving K-Means clustering and snake model, both feature extraction and selection stages using the general approach proposed by WND-CHARM, the classification stage using Random Forest, and the authentication stage using a dynamic threshold maximizing the number of inliers while rejecting outliers.

\paragraph{Acknowledgements}The present work has been developed under the framework of the APIFRESH project (FP7-SME-2008-2/243594), which is co-funded by the European Commission. The authors would like to thank Rafael Redondo, Amelia Gonzalez and Cristina Pardo for their help in the microscope image acquisition.





\bibliographystyle{elsarticle-num}







\end{document}